\documentclass{fairmeta}
\usepackage{amsmath}
\usepackage{enumerate} 
\usepackage{bm}
\usepackage{algorithm}
\usepackage{floatflt}
\usepackage{algpseudocode}
\usepackage{amsfonts}
\usepackage{amsthm}
\usepackage{newtxtt}
\usepackage{algorithm}
\usepackage{colortbl} 
\usepackage{cleveref}
\usepackage{diagbox} 
\usepackage[utf8]{inputenc}
\usepackage{textgreek}
\usepackage{colortbl}
\usepackage{nicematrix}
\usepackage{makecell}
\usepackage{float}
\usepackage{arydshln}
\usepackage[frozencache,cachedir=.]{minted}
\usepackage{caption}
\usepackage{subcaption}
\usepackage{tcolorbox}
\usepackage{amssymb}
\usepackage{xspace}
\usepackage{wrapfig}
\usepackage{adjustbox}
\usepackage{tabularx}
\usepackage{booktabs}
\usepackage{mathtools}
\usepackage{wrapfig}
\usepackage{amssymb}
\usepackage{graphicx}

\usepackage{silence}
\makeatletter
\patchcmd{\wrong@fontshape}{\@gobbletwo}{}{}{}
\makeatother
\definecolor{upColor}{RGB}{17,138,21}
\definecolor{downColor}{RGB}{174,36,67}

\newtheorem{theorem}{Theorem}[]

\newtheorem{remark1}[theorem]{Remark}

\title{Theoretical Foundations of Scaling Law in Familial Models}
\author[]{Huan Song}
\author[]{Qingfei Zhao}
\author[]{Ting Long}
\author[]{Shuyu Tian}
\author[]{Hongjun An}
\author[]{Jiawei Shao}
\author[]{Xuelong Li}
\affiliation[]{Institute of Artificial Intelligence (TeleAI), China Telecom}
\metadata[Keywords]{Familial Models, scaling law, large language models, theoretical analysis, compute-optimal training}

\metadata[Correspondence to]{Xuelong Li (\email{xuelong\_li@ieee.org})}

\begin{document}

\abstract{
Neural scaling laws have become foundational for optimizing large language
model (LLM) training, yet they typically assume a single dense model output.
This limitation effectively overlooks ``Familial Models'', a transformative
paradigm essential for realizing ubiquitous intelligence across heterogeneous
device--edge--cloud hierarchies. Transcending static architectures, Familial
Models integrate early exits with relay-style inference to yield $G$
deployable sub-models from a single shared backbone. In this work, we
theoretically and empirically extend scaling laws to capture this
``one-run, many-models'' paradigm by introducing granularity ($G$) as a
fundamental scaling variable alongside model size ($N$) and training tokens
($D$). To rigorously quantify this relationship, we propose a unified
functional form $L(N,D,G)$ and parameterize it using large-scale empirical
runs. Specifically, we employ a rigorous IsoFLOP experimental design to
strictly isolate architectural impact from computational scale. Across fixed
budgets ($10^{19}$--$10^{21}$ FLOPs), we systematically sweep model sizes
 and granularities  while dynamically adjusting tokens. To further resolve behavior at the level of individual exits, we
introduce a branch scaling law , revealing that
additional upstream branches have negligible impact on performance. Building on these laws, we define an Efficiency Leverage metric ($EL$)
that compares the average loss of Familial Models to that of independent
size-matched dense models under equal FLOPs,  using the fitted scaling
relations, we find that $EL > 1$ across all compute regimes and granularities
considered, with the advantage most pronounced in the low-compute regime.Theoretically, this bridges fixed-compute training with dynamic architectures. Practically, it validates the ``train once, deploy many'' paradigm, demonstrating that deployment flexibility is achievable without compromising the compute-optimality of dense models. 
}

\maketitle

\section{Introduction}
In the landscape of modern Large Language Model (LLM) deployment, diverse applications impose varying constraints on latency and computational cost \citep{kwon2023vllm,hou2025llm,semerikov2025edge}. Practitioners are no longer satisfied with a single fixed model; instead, there is a pressing need for a flexible suite of models capable of spanning multiple cost tiers \citep{chen2023frugalgpt,huang2025thriftllm,park2024any}. To address this, ``Familial Models'' have emerged as a transformative solution \citep{an2506ai}. Going beyond standard early-exit architectures \citep{teerapittayanon2016branchynet}, the proposed approach synergistically integrates Early Exiting with Scalable Branches (EESB) and Hierarchical Principal Component Decomposition (HPCD). Specifically, instead of merely attaching prediction heads to intermediate layers, lightweight, decomposable branch networks \citep{houlsby2019parameter} are constructed to allow for fine-grained parameter tuning via low-rank matrix approximation \citep{hu2022lora}. This architecture enables a single training run to produce $G$ deployable sub-models (where $G$ denotes granularity) that share a unified backbone and aligned hidden features \citep{kusupati2022matryoshka}. This structural consistency not only offers a continuous spectrum of depth-cost trade-offs but also inherently supports relay-style cooperative inference across heterogeneous devices\citep{kang2017neurosurgeon} without additional middleware, thereby significantly enhancing the flexibility and efficiency of the ``train once, deploy many'' paradigm.

To guide efficient model training and resource allocation, the field relies heavily on Neural Scaling Law. The foundational era began with \citet{kaplan2020scaling}, who characterized test loss as a predictable power-law function of model size ($N$), dataset size ($D$), and compute budget ($C$). This paradigm was significantly refined by \citep{hoffmann2022training} through IsoFLOP analysis, establishing the ``Chinchilla'' scaling law which advocates for proportional scaling of parameters and data ($N \propto D$) to maximize efficiency. Recent rigorous replications have further solidified this foundation; despite identifying methodological flaws in the original Chinchilla study—such as optimizer early stopping and parameter rounding errors—researchers reaffirmed the validity of the compute-optimal frontier with corrected, statistically robust confidence intervals \citep{pearce2024reconciling,porian2024resolving}.

As the field evolves, scaling laws are expanding beyond dense models to specialized, efficient architectures. For instance, recent work on Mixture-of-Experts (MoE) introduced ``Efficiency Leverage'' (EL) to quantify the computational advantage over dense models \citep{tian2025towards}. This research revealed that efficiency is governed by distinct architectural factors: EL scales as a power law with the activation ratio (sparsity) and exhibits a non-linear ``U-shaped'' sensitivity to expert granularity, with advantages amplifying significantly at larger compute budgets \citep{tian2025towards,krajewski2024scaling}. These advances reflect a broader paradigm shift: as the community navigates potential saturation in pure scaling and explores new frontiers like post-training scaling and data quality, the focus is moving toward architecture-aware laws that ensure precise resource optimization.

However, existing scaling laws are inherently built upon a ``one-run, one-model'' paradigm \citep{Yuan2025TaskOrientedFC}, characterizing loss solely as a function of $N$ and $D$ for a single output. This perspective fails to capture the unique ``one-to-many'' dynamics of Familial models training, where the outcome is not a solitary model but a set of $G$ interdependent sub-models derived from a single optimization process. Traditional laws overlook the architectural dimension of ``Granularity'' ($G$, the number of exit points), and thus cannot quantify the potential interference or synergy between exits, nor predict the performance cost of making a model family ``finer-grained.''
To bridge this theoretical gap, we propose a unified scaling framework that explicitly incorporates Granularity ($G$) as a fundamental scaling variable alongside $N$ and $D$. Drawing inspiration from the architectural deconstruction approach of  \citet{tian2025towards}, our methodology proceeds as follows:

\begin{enumerate}
    \item \textbf{Familial Models Scaling Law:} To quantify the modulatory effect of granularity, we adopt the formulation:
\begin{equation}
    L(N,D,G) = \left( E + \frac{A}{N^{\alpha}} + \frac{B}{D^{\beta}} \right) \cdot G^{\gamma},
\end{equation}
This multiplicative structure isolates the granularity penalty $G^{\gamma}$ from the standard power-law decay, effectively interpreting $\gamma$ as the marginal ``tax'' imposed on the backbone for supporting multiple independent operating points.
By parameterizing the unified functional form with data from our rigorous IsoFLOP experiments, we derive the \textit{Familial Models Scaling Law}. The fitted law is quantitatively established as:
\begin{equation}
    L(N,D,G) = \left( 1.0059 + \frac{403.4289}{N^{0.2982}} + \frac{2980.958}{D^{0.3412}} \right) \cdot G^{0.0333},
\end{equation}
In this equation, the irreducible loss $E=1.0059$ represents the theoretical performance limit, while the small exponent $\gamma \approx 0.0333$ empirically confirms that the architectural overhead for supporting $G$ exits is minimal, following a gentle multiplicative scaling rule.
    \item \textbf{Branch Scaling Law in Familial Models:} To quantitatively analyze the performance gap between branch models in
Familial Models and size-matched dense models, we propose the following
branch-level scaling relation ($P$ denotes the number of branch points preceding the
given branch within the family, and $Q$ denotes the number of branch points
succeeding it):
\begin{equation}
    L(P,Q,D) = L_{\mathrm{dense}} + (\alpha\cdot P + \beta\cdot Q)\cdot\left(\frac{D_d}{D}\right)^{a},
\end{equation}
This formulation explicitly disentangles the penalties on loss induced by
additional branch points before and after the exit under consideration, with
$\alpha$ and $\beta$ capturing their respective contributions. By fitting this
functional form to more than one hundred experimental configurations, we obtain
a branch-level scaling law for Familial Models that quantitatively characterizes
the performance gap between familial branches and size-matched dense models.
The fitted law is quantitatively established as:
\begin{equation}
L(P, Q, D) = L_{\mathrm{dense}} 
  + \bigl(1\times 10^{-3} \, \cdot P + 0.0397 \, \cdot Q\bigr)
    \cdot\left(\frac{2.75\times 10^{6}}{D}\right)^{0.5734},
\end{equation}
This scaling relation further implies that adding additional branch points \emph{before} a given exit has only a negligible effect on its performance, indicating that Familial Models can introduce multiple sub-models of different sizes with virtually no degradation in the quality of the underlying backbone model.
    \item \textbf{Efficiency Leverage:} Motivated by the efficiency leverage metric proposed for MoE models in \cite{tian2025towards}, we define an analogous Efficiency Leverage (EL) for Familial Models to quantify their compute advantage over dense models. Specifically, EL is defined as the ratio of the average loss of a dense baseline to that of a Familial Model under a matched compute budget:
\begin{equation}
EL(\mathcal{X}_{\mathrm{Fam}} \mid \mathcal{X}_{\mathrm{Dense}})
  = \frac{L_{\mathrm{Dense}}}{L_{\mathrm{Fam}}},
\end{equation}
The results show that $EL$ remains strictly greater than $1$ across all compute
budgets, indicating that Familial Models achieve lower loss than dense models under matched FLOPs budgets. Moreover, this advantage is amplified in the low-compute regime,
where the Efficiency Leverage attains its highest values.
    \item \textbf{Compute-Matched Parameterization:}
To ensure the scaling coefficients accurately reflect the architectural trade-offs of Familial Models, we parameterize the scaling law using a dataset derived from strict IsoFLOP (constant compute) constraints. By systematically varying model size $N$ and granularity $G$ within fixed compute budgets ($10^{19}$--$10^{21}$ FLOPs), we generate a high-fidelity observation set. This design effectively decouples the marginal cost of granularity from computational scale, allowing for a precise isolation of the granularity exponent $\gamma$ independent of the fitting algorithm employed.
\end{enumerate}

This work pioneers the theoretical formalization of the ``Familial Models'' paradigm, establishing the first unified scaling law that explicitly incorporates Granularity ($G$) as a fundamental dimension alongside model size ($N$) and data ($D$). Unlike prior studies limited to static dense models, we quantify the ``cost of flexibility'' by accurately modeling the three-dimensional loss surface. A critical discovery from our results is that the granularity exponent $\gamma$ is extremely small ($\approx 0.033$), quantitatively proving that the architectural penalty for supporting multiple exit points is negligible. Beyond this global law, we introduce a branch-level scaling law for Familial
Models that quantifies the performance gap between individual branches and
size-matched dense models. The fitted coefficients show that additional
branch points \emph{before} a given exit have only a very minor impact on its
loss, indicating that the familial architecture can produce multiple
sub-models of different sizes from a single training run with virtually no
degradation in backbone performance. Finally, we define an Efficiency Leverage
metric $EL$ and, using the fitted familial scaling law, compute its variation
across different FLOPs budgets. The resulting curves confirm that $EL$ exceeds
$1$ for all compute regimes considered, implying that Familial Models are
consistently more compute-efficient than dense models, with the advantage most
pronounced in the low-compute regime.
\section{Preliminaries}

\subsection{Familial Models Architectures and Granularity Formulation}

We adopt a conventional dense Transformer equipped with a single final prediction head as the baseline, which corresponds to the special case $G = 1$. Within each compute-budget group, we fix the total training compute (FLOPs; ranging from $10^{19}$ to $10^{21}$) and, for each architectural configuration, derive the corresponding training-token budget $D$ implied by the fixed compute constraint. This ensures fair, compute-matched comparisons across models within the same group.

Familial Models is built upon a shared backbone and augments it with multiple early-exit prediction heads placed at selected intermediate layers. This design allows a single trained trunk to produce multiple deployable sub-models with different effective depths and inference costs. Formally, let the trunk contain \textit{\textit{L}} transformer layers. We attach exit heads at a set of intermediate layers\{\textit{\textit{l}}\textsubscript{\textsubscript{1}},...,\textit{\textit{l}}\textit{\textsubscript{\textsubscript{\textit{G-1}}}}\} and also retain the standard output at the final layer \textit{\textit{L. }}This yields a total of \textit{\textit{G }}usable exits (including the final exit). We define \textit{\textit{G }}as the granularity factor, where a larger \textit{\textit{G }}corresponds to more available operating points (i.e., more depth/cost tiers) and thus finer deployment granularity. Training typically optimizes all exits jointly by minimizing a weighted sum of exit-specific language-modeling losses:

\begin{equation}
\mathcal{L}_{\text{family}} = \sum_{g=1}^{G} w_g \mathcal{L}_g,
\end{equation}
Where $\mathcal{L}_{g}$ denotes the language-modeling loss at exit $g$, and $w_{g}$ is the corresponding weight. In our implementation, we assign equal weights to all exits (i.e., $w_{g} = 1/G$), such that the total Familial Models loss is defined as the arithmetic mean of the individual losses across all exits. Within each experimental group, we keep the exit-weighting scheme and training protocol fixed, so that the primary independent variables for scaling analysis are $(N, D, G)$.

\subsection{Extending Scaling Law to Familial Models}
Modern deployment environments are rarely uniform; they often demand a versatile suite of models spanning a wide range of latency and cost tiers to adapt to varying hardware constraints (e.g., server-side vs.\ on-device) and dynamic query complexities. Relying on a single fixed operating point is inefficient, yet training independent models for each desired tier incurs a prohibitive computational cost that scales linearly with the number of models. Familial Models offer an elegant solution to this dilemma by training a shared Transformer trunk equipped with multiple intermediate exits. In this architecture, a single training run yields $G$ deployable sub-models, each representing a distinct effective depth and inference cost. Here, the granularity $G$ serves as a critical architectural hyperparameter, directly quantifying the density of valid operating points available from a single backbone. It effectively measures the ``deployment resolution'' of the model family---a higher $G$ implies a finer-grained ability to trade off accuracy for speed without the need for retraining.

Classical scaling laws (e.g., Kaplan et al., 2020; Hoffmann et al., 2022) have provided robust guidelines for predicting how loss varies with parameter count $N$ and training tokens $D$. However, these frameworks operate under the assumption of distinct, independently trained models, failing to account for the internal dependencies and weight sharing inherent in multi-exit architectures\footnote{Unlike independent models, sub-models in a family architecture share the majority of their parameters. This creates a multi-objective optimization landscape where gradients from deeper exits can regularize or potentially interfere with shallower ones, a dynamic not captured by traditional scaling laws for dense models.}. In the Familial models setting, the training outcome is not a solitary scalar loss, but a trajectory of losses across $G$ entangled sub-models derived from the same optimization process. To capture this ``one-run, many-models'' paradigm within a unified theoretical framework, we explicitly incorporate $G$ as a third fundamental scaling dimension alongside $N$ and $D$. By studying the joint scaling function $L(N, D, G)$, we aim to rigorously quantify the marginal cost of granularity---determining whether the architectural overhead of supporting multiple exits alters the fundamental compute-optimal frontier established for dense models.

\section{Scaling Law with Granularity}
Next, we define the functional form of the familial models scaling law and use it to outline our objectives and experimental roadmap. 
\subsection{Proposed functional form}
Following empirical scaling-law literature—particularly compute-optimal scaling analyses (e.g., Hoffmann et al.)---we model pretraining loss as a smooth, monotone function of model size and data scale, exhibiting diminishing returns and approaching a non-zero irreducible floor. In the dense setting, this behavior is well captured by an additive decomposition in which loss approaches an irreducible term and decays as power laws in $N$ and $D$. To extend this perspective to Familial Models, we introduce granularity ($G$) and propose a unified scaling law, utilizing the methodology of Hoffmann et al. to fit our extended parametric form to strictly compute-matched training runs:

\begin{equation}
L(N, D, G) = \left( E + \frac{A}{N^\alpha} + \frac{B}{D^\beta} \right) \cdot G^\gamma,\tag{1}
\end{equation}
where $E$ is the irreducible loss floor as $N,D\to\infty$, $A$ and $B$ are positive scale coefficients, and $\alpha,\beta>0$ govern the rates of power-law improvement from increasing model size and data scale. The term $G^{\gamma}$ captures the multiplicative effect of granularity on the learnable component of the loss. This formulation reduces to the standard $(N,D)$ scaling law when $G=1$, while remaining compact, interpretable, and straightforward to fit from empirical runs.

\subsection{Fitting Procedure}
A standard decomposition-based fitting procedure is adopted in the log domain \citep{hoffmann2022training} , combining robust regression with multi-start initialization to ensure numerical stability and reliable parameter estimation. 
\begin{itemize} \item \textbf{\textbf{Log-domain decomposition with LSE:}} To facilitate gradient-based optimization and ensure numerical stability, the standard scaling law formulation is reformulated as: 
\begin{equation}
L(N, D, G) = \left( E + A N^{-\alpha} + B D^{-\beta} \right) G^\gamma,
\end{equation}
Ensuring positivity and optimization stability, the coefficients are parameterized by:\[
E = \exp(e), \quad A = \exp(a), \quad B = \exp(b),\tag{8}
\]
For a specific run \textit{\textit{i}}, the predicted log value $\log \hat{L}_i$ is calculated using these exponential terms:
\[
\log \hat{L}_i = \log \left( \exp(e) + \exp(a - \alpha \log N_i) + \exp(b - \beta \log D_i) \right) + \gamma \log G_i,\tag{9}
\]
Specifically, we implement the log of the positive sum via a log-sum-exp operator defined as:
\[
\text{LSE}(x, y, z) = \log \left( \exp(x) + \exp(y) + \exp(z) \right),\tag{10}
\]
This results in the final formulation:  
\[
\log \hat{L}_i = \text{LSE}(e, a - \alpha \log N_i, b - \beta \log D_i) + \gamma \log G_i,\tag{11}
\]
\item \textbf{\textbf{Robust objective: Huber loss on log-residuals:}} The discrepancy between the model's prediction and the observed data is quantified by the log-residual $r_i$, which is defined as: 
\[
r_i = \log \hat{L}_i - \log L_i,\tag{12}
\]
The model is fitted by minimizing the sum of Huber losses: 
\[
\min_{a, b, e, \alpha, \beta, \gamma} \sum_{i \in R} \text{Huber}_\delta (r_i),
\]
We use $\delta = 10^{-3}$ for Huber robustness\footnote{Training runs occasionally exhibit transient loss spikes or instabilities, particularly in early phases. Unlike Mean Squared Error (MSE), which heavily penalizes these outliers and can skew the fitted curve, Huber loss transitions to linear scaling for large residuals, thereby effectively ignoring these non-representative data points.}. Empirically, larger $\delta$ tends to overfit lower-compute regimes and predict held-out larger-compute runs poorly, while $\delta < 10^{-3}$ does not materially change the resulting predictions—consistent with robust behavior.
\item  \textbf{\textbf{Optimization: L-BFGS with grid initialization:}} As the objective function is non-convex, L-BFGS is employed to locate high-quality local minima. To further reduce sensitivity to initialization, optimization is initialized from a grid of starting points, which improves stability and consistency of the fitted solutions. The solution achieving the lowest final objective value is selected. In our experiments, the optimal solution does not occur at the boundary of the initialization grid, indicating that the resulting fit is unlikely to be an artifact of the chosen grid limits. 
\[
\alpha \in \{0, 0.5, \dots, 2\}, \quad \beta \in \{0, 0.5, \dots, 2\}, \quad e \in \{-1, -0.5, \dots, 1\},
\]
\[
a \in \{0.5, \dots, 25\}, \quad b \in \{0.5, \dots, 25\}, \quad\gamma \in \{0, 0.5, \dots, 2\},
\]

\end{itemize}
\subsection{Experimental Design}
We conduct a series of experimental groups designed to support reliable scaling-law estimation under controlled compute conditions. In each group, we fix the overall training compute budget (FLOPs). For every configuration (dense models and Familial Models variants), we derive the corresponding training-token budget \textit{\textit{D}} implied by the fixed compute constraint, accounting for the fact that different exit layouts may alter the effective per-token training cost\footnote{Specifically, the forward and backward pass computations of the additional exit heads are included in the total FLOPs count. Therefore, for a fixed compute budget, increasing the granularity $G$ (i.e., adding more heads) results in a slightly higher per-token cost, necessitating a compensatory reduction in the training token count $D$ compared to a dense baseline.}.

Under the same compute budget, we then sweep across a range of parameter scales \textit{\textit{N}} and evaluate multiple Familial Models design with different granularity settings \textit{\textit{G }}implemented by varying both the number and placement of intermediate exits. Each training run yields an observation (\textit{\textit{N}}\textit{\textsubscript{\textsubscript{\textit{i}}}}\textit{\textit{,D}}\textit{\textsubscript{\textsubscript{\textit{i}}}}\textit{\textit{,G}}\textit{\textsubscript{\textsubscript{\textit{i}}}}\textit{\textit{,L}}\textit{\textsubscript{\textsubscript{\textit{i}}}}). The union of runs across all groups forms the dataset used to fit the proposed scaling law. For clarity, we present one representative experimental group to illustrate concrete architecture choices and exit placements. Each experimental group contains both a dense model baseline and a Familial Model variant; the Dense Model experimental setup is summarized in Table \ref{tab:dense_hparams}, and the Familial Models experimental setup is summarized in Table \ref{tab:family_model_hparams}. Unless otherwise noted, all reported scaling-law parameters are obtained by fitting to the full set of runs across all groups.

\begin{table*}[htb]
    \centering
    \begin{NiceTabular}{lccccc}
        \toprule
        \textbf{Dense} & \textbf{d\_model} & \textbf{ffn\_size} & \textbf{num\_attention\_heads} & \textbf{$\text{n\_layers}$}\\
        \midrule
        1B & 1536 & 4608 & 12 & 19 \\
        2B & 2048 & 6144 & 16 & 27 \\
        3B & 2304 & 6912 & 18 & 36 \\
        4B & 2560 & 7680 & 20 & 41 \\
        \bottomrule
    \end{NiceTabular}
    \captionsetup{justification=centering} 
    \caption{Architectural hyperparameters of the dense transformer baselines used in our experiments.}
    \label{tab:dense_hparams}
\end{table*}

\begin{table*}[htb]
    \centering
    \begin{NiceTabular}{lccccc}
        \toprule
        \textbf{Family} & \textbf{d\_model} & \textbf{ffn\_size} & \textbf{num\_attention\_heads} & \textbf{$\text{n\_layers}$} & \textbf{exit\_layer} \\
        \midrule
        2B & 2048 & 6144 & 16 & 27 & 10 \\
        3B & 2304 & 6912 & 18 & 36 & 6, 20 \\
        4B & 2560 & 7680 & 20 & 41 & 4, 16, 18 \\
        \bottomrule
    \end{NiceTabular}
    \captionsetup{justification=centering} 
    \caption{Architectural hyperparameters of the Familial
Models variants and their exit configurations.}
    \label{tab:family_model_hparams}
\end{table*}

\section{Results}
\subsection{Analysis  and  Visualization  of Fitted  Familial Models Scaling Law  }

For the representative experimental group, the fitted scaling relation takes the form:
\begin{equation}
L(N, D, G) = \left( 1.0059 + \frac{403.4289}{N^{0.2982}} + \frac{2980.058}{D^{0.3412}} \right) \cdot G^{0.0333},\tag{2}
\end{equation}

\begin{figure}[H]
\centering
\includegraphics[width=0.75\linewidth]{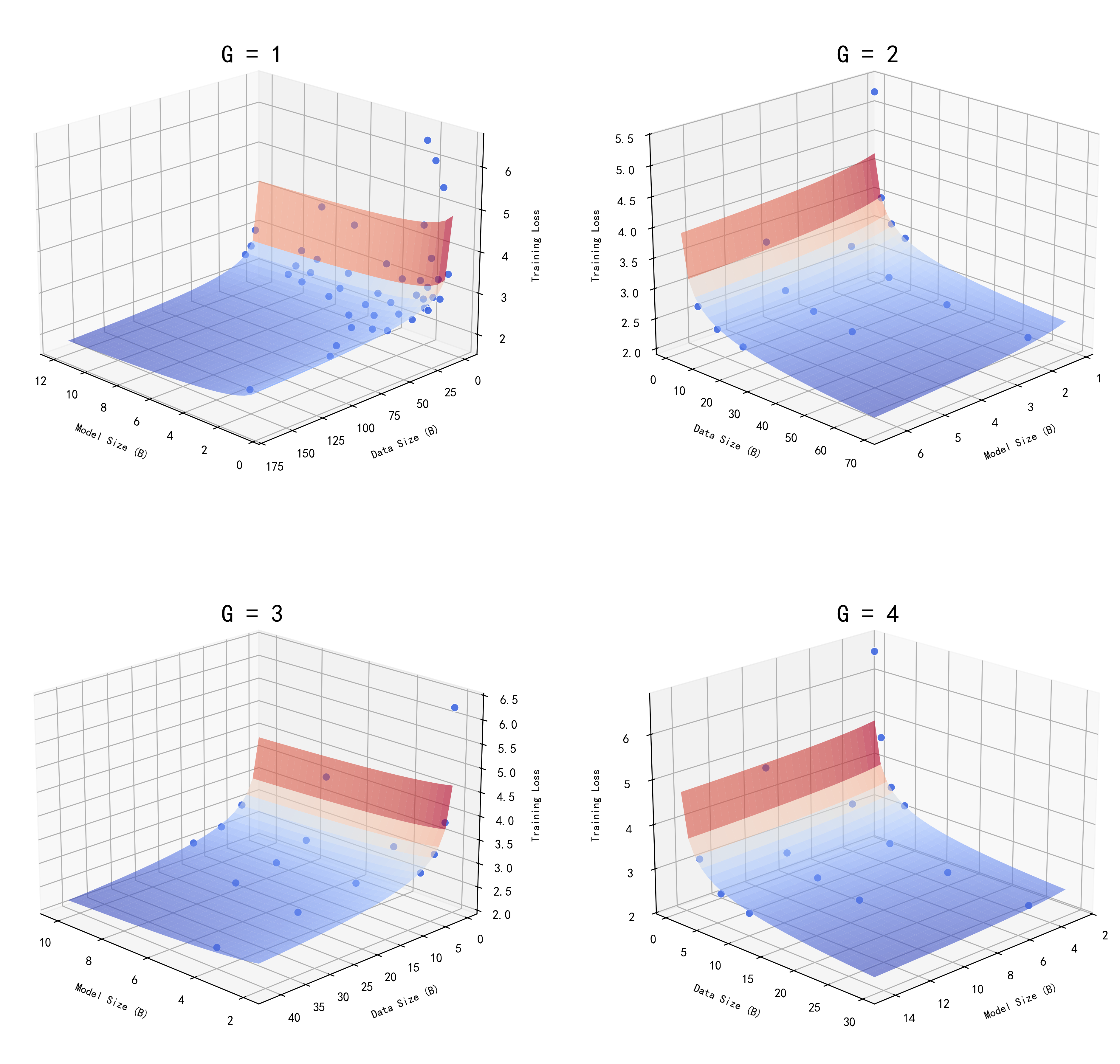}
\caption{\textbf{Three-dimensional visualizations of the fitted Familial Models scaling law
for different granularity levels $G$.} The four panels correspond to
$G \in \{1,2,3,4\}$. In each panel, the horizontal axes represent the model
size $N$ and training token count $D$ (both in log scale), while the vertical
axis shows the fitted loss $L$; blue markers denote individual training runs
and the shaded surface indicates the fitted loss surface. Across all values of
$G$, the loss decreases smoothly as model and data scale increase, while the
slight vertical shift of the surfaces with increasing $G$ reflects the minimal
multiplicative penalty implied by the small exponent $\gamma \approx 0.0333$,
indicating that Familial Models retain near-compute-optimal scaling behavior.}
\label{fig:family_model_scaling_law}
\end{figure}

To facilitate interpretation, we visualize the learned relationship as a set of
three-dimensional loss surfaces indexed by granularity $G$. In
Figure~\ref{fig:family_model_scaling_law}, each panel corresponds to a fixed
$G$ and plots the fitted loss $L(N,D \mid G)$ as a function of model scale $N$
and training token count $D$. The surfaces exhibit the characteristic smooth
decay in loss as both model size and data scale increase. Comparing panels
across different $G$, we observe only a mild upward shift of the surfaces,
consistent with the small but systematic multiplicative penalty captured by the
exponent $\gamma = 0.0333$.

\subsection{Efficiency Frontier}
\begin{figure}[H]
\centering
\includegraphics[width=0.75\linewidth]{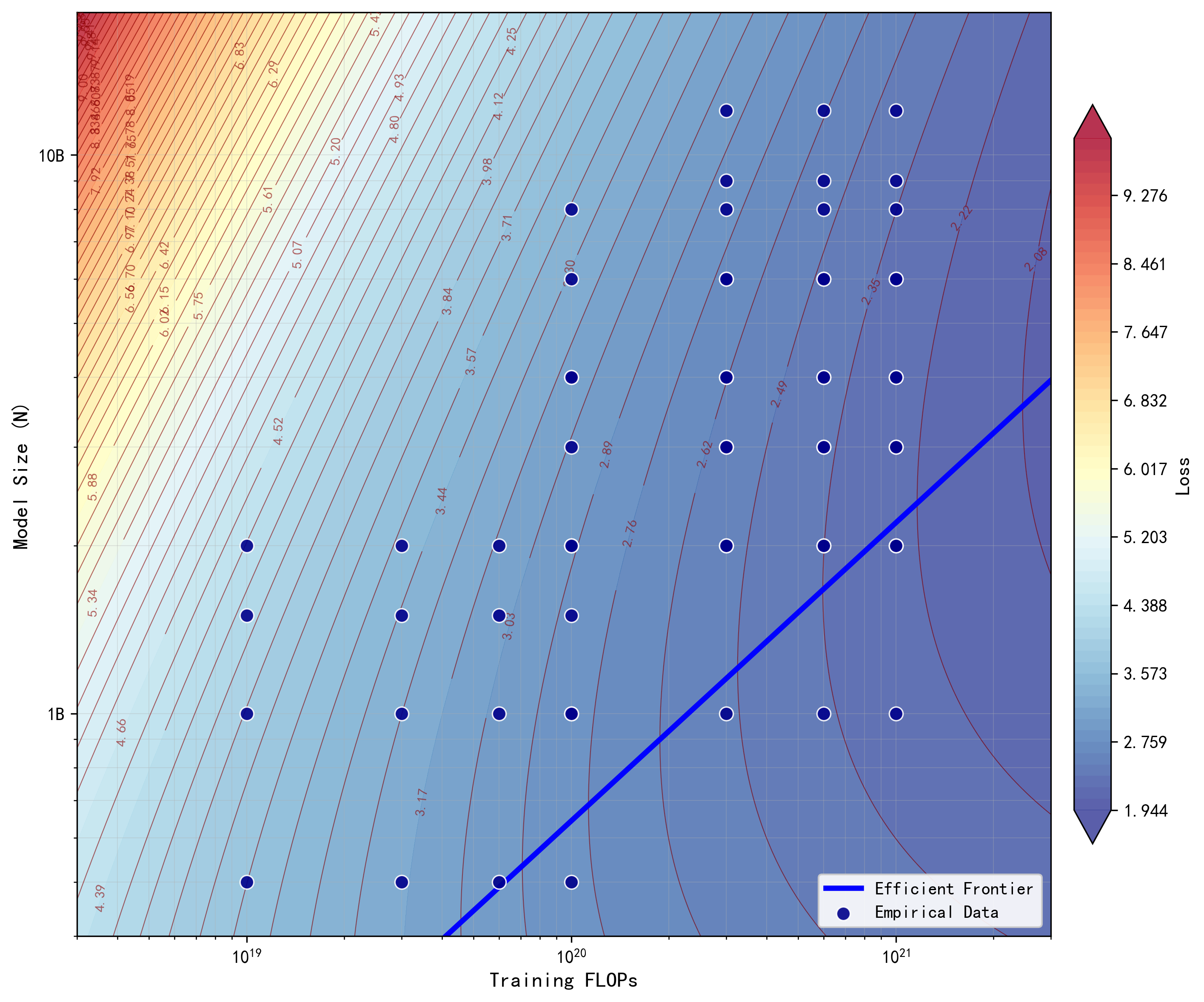}
\caption{\textbf{Compute-efficient frontier.} The plot displays the isoloss contours and the implied efficiency frontier (blue line) across varying FLOPs budgets. This line represents the compute-optimal allocation of model size for a given training budget.}
\label{fig:compute_efficient_frontier_g3}
\end{figure}

Figure \ref{fig:compute_efficient_frontier_g3} illustrates the efficiency frontier, providing a quantitative framework to guide the training and design of Familial Models. By mapping the relationship between model size \textit{N} and data scale \textit{D}, this frontier facilitates a principled approach to the ``one-run, many-models'' paradigm.

\subsection{Branch Scaling in Familial Models}
\begin{figure}[H]
\centering
\includegraphics[width=1\linewidth]{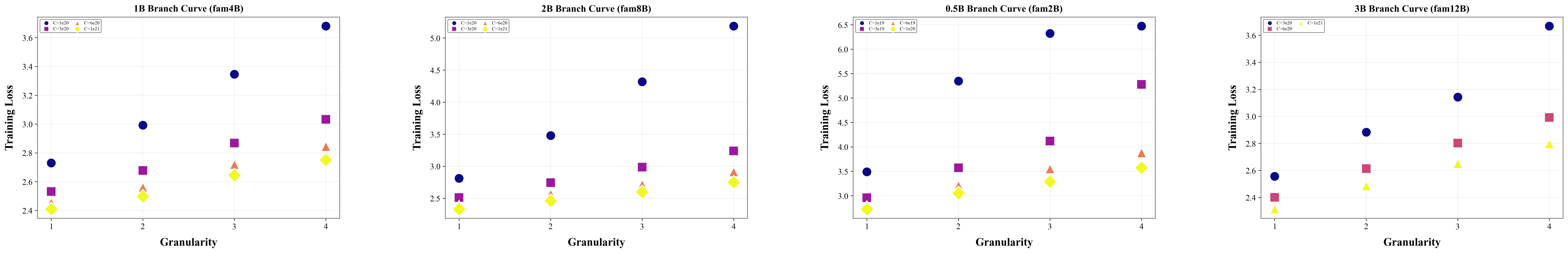}
\caption{\textbf{Loss at the first branching point: Familial versus Dense models.} Training loss of the first-branch models in each Familial backbone (0.5B in
\texttt{fam2B}, 1B in \texttt{fam4B}, 2B in \texttt{fam8B}, and 3B in
\texttt{fam12B}) compared with independently trained dense model of matching
parameter count, plotted as a function of granularity $G$. }
\label{fig:branch_first_vs_dense}
\end{figure}

\begin{figure}[H]
\centering
\includegraphics[width=1\linewidth]{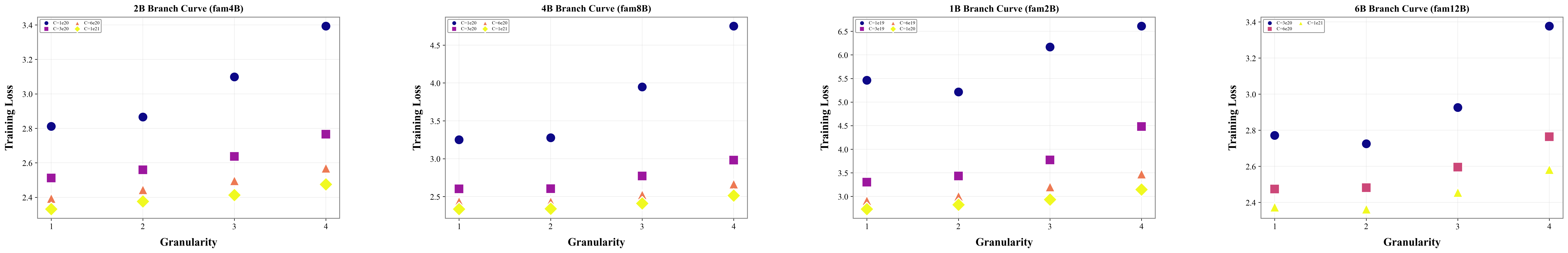}
\caption{\textbf{Loss at the second branching point: Familial versus Dense models.} Training loss of the second-branch model in each Familial backbone
(1B in \texttt{fam2B}, 2B in \texttt{fam4B}, 4B in \texttt{fam8B}, and 6B in
\texttt{fam12B}) compared with independently trained dense models of matching
parameter count, plotted as a function of granularity $G$. }
\label{fig:branch_second_vs_dense}
\end{figure}

We investigate the performance differences between branch models in Familial Models and their dense counterparts under matched parameter counts and compute budgets. Specifically, for the \texttt{fam4B} family, we examine the 1B and 2B branch models against independently trained dense models of the same parameter size across four distinct FLOP budgets. The same experimental protocol is applied to the \texttt{fam2B} and \texttt{fam12B} families. Figure~\ref{fig:branch_first_vs_dense} reports the loss curves at the first branching point in each family compared to dense models of identical size, while Figure~\ref{fig:branch_second_vs_dense} presents the corresponding comparison for the second branching point.

From Figure~\ref{fig:branch_first_vs_dense}, we observe that as the granularity $G$ increases, the training loss increases monotonically, exhibiting an approximately linear relationship with $G$. In contrast, Figure~\ref{fig:branch_second_vs_dense} reveals a more nuanced pattern: increasing $G$ from 1 to 2 incurs only a marginal loss penalty, whereas for $G \geq 2$, the relationship between granularity and loss returns to a near-linear trend.

This discrepancy arises from how additional branches are positioned relative to the branch under consideration. In Figure~\ref{fig:branch_first_vs_dense}, increasing $G$ corresponds to adding branches \emph{after} the evaluated branch point, whereas in Figure~\ref{fig:branch_second_vs_dense}, the change from $G = 1$ to $G = 2$ is realized by inserting an additional branch \emph{before} the evaluated branch point; only when $G$ increases from 2 to 4 are new branches appended downstream. Consequently, adding branches downstream of a given branch has a significantly larger impact on its loss than adding branches upstream, and this impact is approximately linear with respect to the number of downstream additions. Moreover, as the compute budget increases, the slope of this linear relationship decreases, implying that with more abundant compute, the loss gap between different granularities diminishes. This behavior is consistent with the training dynamics: gradients from downstream branches propagate through and update all parameters of the evaluated branch's backbone, whereas gradients from upstream branches affect only the shared earlier layers (a subset of the parameters).

Building on the above empirical results, we propose a scaling law for branch
models in Familial Models that characterizes the additional loss incurred when
introducing extra branch points compared to a dense baseline:
\[
L(P, Q, D) = L_{\mathrm{dense}} + (\alpha\cdot P + \beta \cdot Q)\cdot\left(\frac{D_d}{D}\right)^{a}\tag{3},
\]
where $L(P, Q, D)$ denotes the loss of the considered branch model and
$L_{\mathrm{dense}}$ is the loss of a dense model with matched parameter count
and compute. The term $P$ denotes the number of branch points preceding the
given branch within the family, and $Q$ denotes the number of branch points
following it. Since, during training, downstream branches typically exert a
stronger influence on the performance of upstream branches than the reverse, we
assign distinct penalty weights $\alpha$ and $\beta$ to $P$ and $Q$,
respectively. The factor $\left(D_d / D\right)^{a}$ further captures the
dependence of this penalty on the compute budget through the ratio between token budget $D$ .

\begin{figure}[H]
\centering
\includegraphics[width=1.0\linewidth]{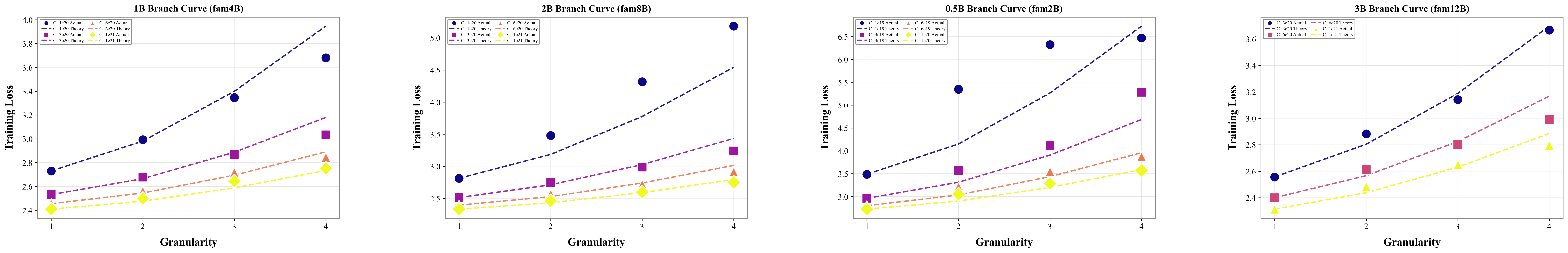}
\caption{\textbf{Loss at the first branching point: empirical vs.\ scaling-law
predictions.} The training loss of the
first-branch models in each Familial backbone (0.5B in \texttt{fam2B}, 1B in
\texttt{fam4B}, 2B in \texttt{fam8B}, and 3B in \texttt{fam12B}) is plotted as
a function of granularity $G$ under several IsoFLOP budgets $C$. Solid markers
denote empirical measurements, while dashed curves show the corresponding
predictions from the proposed branch scaling law.}
\label{fig:branch_first_vs_theory}
\end{figure}

\begin{figure}[H]
\centering
\includegraphics[width=1.0\linewidth]{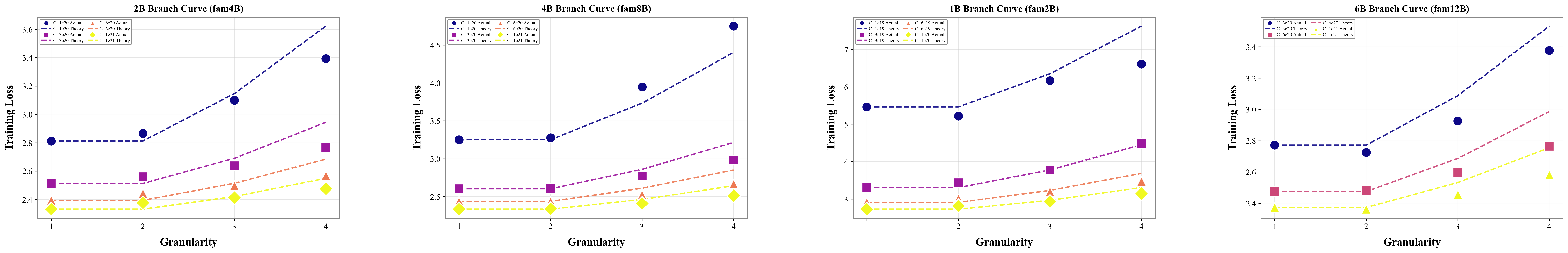}
\caption{\textbf{Loss at the second branching point: empirical vs.\ scaling-law
predictions.} The training loss of the
second-branch models in each Familial backbone (1B in \texttt{fam2B}, 2B in
\texttt{fam4B}, 4B in \texttt{fam8B}, and 6B in \texttt{fam12B}) is plotted as
a function of granularity $G$ under several IsoFLOP budgets $C$. Solid markers
denote empirical measurements, while dashed curves show the corresponding
predictions from the proposed branch scaling law.}
\label{fig:branch_second_vs_theory}
\end{figure}

Using the same fitting procedure as for the preceding scaling laws, we fit the
branch model loss on over 100 experimental configurations and obtain(the corresponding fitted curves based on the above
experimental data are shown in Figure~\ref{fig:branch_first_vs_theory} and Figure~\ref{fig:branch_second_vs_theory}):
\[
L(P, Q, D) = L_{\mathrm{dense}} 
  + \bigl(1\times 10^{-3} \, \cdot P + 0.0397 \, \cdot Q\bigr)
    \cdot\left(\frac{2.75\times 10^{6}}{D}\right)^{0.5734},\tag{4}
\]

The fitted coefficients show that additional branch points \emph{before} a
given branch (captured by $P$) have a much smaller impact on its loss than
additional branch points \emph{after} it (captured by $Q$).
Moreover, this scaling law implies that, under a fixed compute budget, the
largest branch in a Familial Models can match the performance of a dense model
of the same size to within a negligible margin, while simultaneously producing
multiple smaller sub-models. This highlights a key advantage of Familial
Models over traditional dense architectures: they preserve near–compute-optimal
performance at the largest scale while providing a rich set of intermediate
operating points for deployment.

\subsection{Efficiency Leverage}
To quantitatively assess the advantage of Familial Models over dense models, we
introduce the \emph{Efficiency Leverage} (EL), defined as the ratio between the
average loss of multiple dense models and the average loss across all branch
points of a corresponding Familial Models under the same compute budget:
\[
EL(\mathcal{X}_{\mathrm{Fam}} \mid \mathcal{X}_{\mathrm{Dense}})
  = \frac{L_{\mathrm{Dense}}}{L_{\mathrm{Fam}}}\tag{5},
\]

\begin{figure}[H]
\centering
\includegraphics[width=0.5\linewidth]{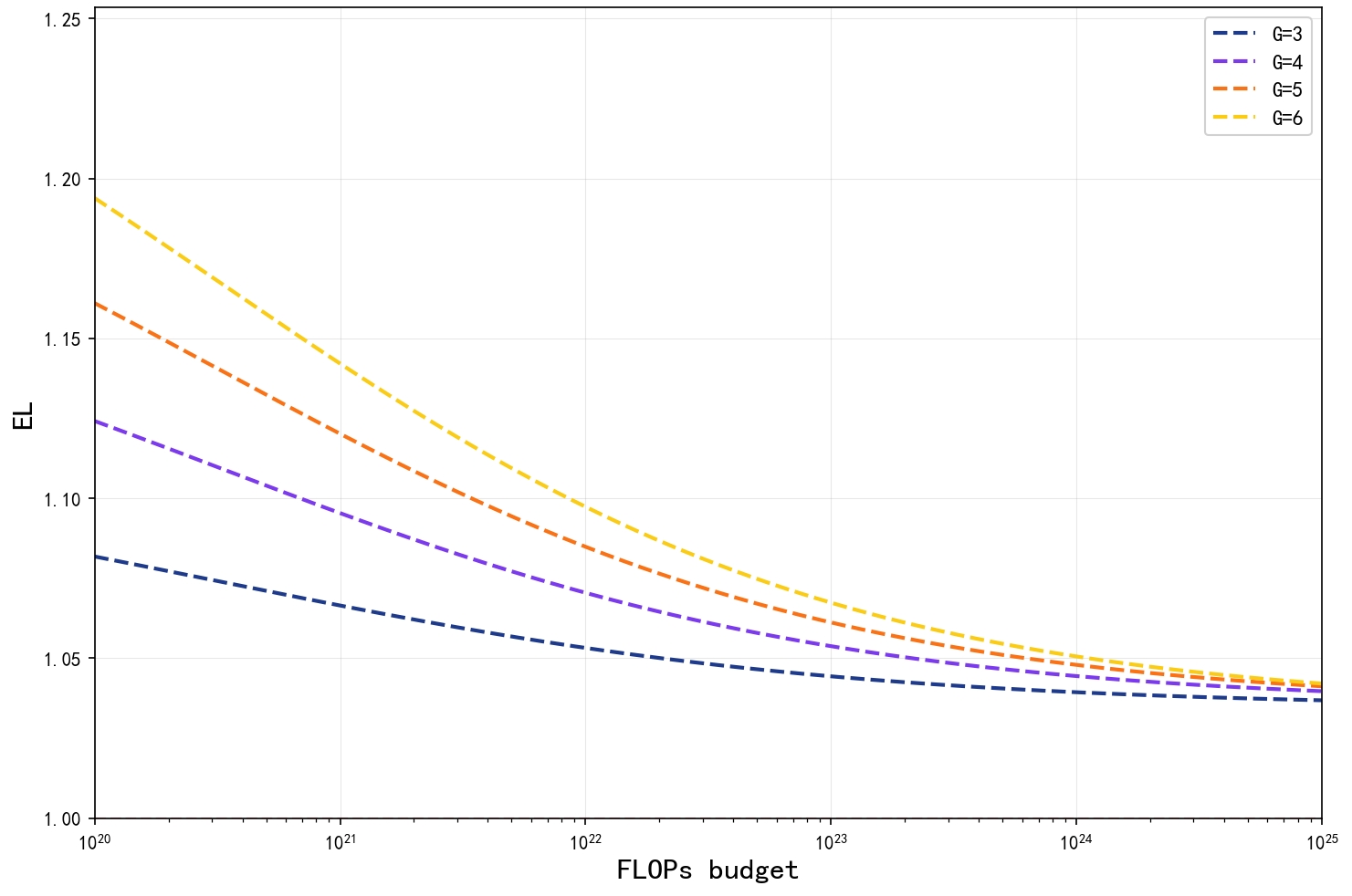}
\caption{\textbf{Efficiency leverage vs.\ FLOPs budget.}
Efficiency leverage $EL(\mathcal{X}_{\mathrm{Fam}} \mid \mathcal{X}_{\mathrm{Dense}})$
computed from the fitted Familial Models scaling law, plotted as a function of
the total training FLOPs budget for different granularities
$G \in \{3,4,5,6\}$.}
\label{fig:EL}
\end{figure}

Based on the fitted Familial Models scaling law, we compute $EL$ as a function
of training FLOPs for different granularities, and plot the resulting curves in
Figure~\ref{fig:EL}. The curves show that $EL$ remains consistently greater
than $1$, indicating that Familial Models achieve lower loss than their dense
counterparts at matched compute. Moreover, for a given FLOPs budget, $EL$
tends to increase with granularity $G$, indicating that architectures
with more exits derive greater relative benefit from the Familial design.
This Efficiency Leverage is most pronounced in the low-compute regime, where
the shared backbone enables Familial Models to make more effective use
of limited training resources.

\section{Discussion}

The establishment of the Familial Models scaling law $L(N,D,G)$, together with
the proposed branch scaling law $L(P,Q,D)$ for Familial Models, not only optimizes the
training of Familial Models but also provides a theoretical foundation for
a wide range of downstream applications that require dynamic resource
adaptation at both the model and branch levels.

\textbf{Empirical Validation of Granularity Efficiency.}
Our rigorous fitting results for the Familial Models scaling law $L(N, D, G)$
reveal that the granularity exponent $\gamma$ is extremely small, indicating
that under matched model size ($N$) and data scale ($D$), increasing
granularity $G$---i.e., adding more exit layers so that a single training run
yields more deployable sub-models---induces only a very mild multiplicative
change in loss. In practical terms, the fitted factor $G^{\gamma}$ stays close
to $1$ over a wide range of $G$, meaning that the loss surface is only weakly
sensitive to the number of exits. Complementing this global law, our
branch-level scaling analysis further quantifies how the loss of an individual
exit depends on the number and placement of additional branch points: extra
branches \emph{before} a given branch have only a negligible effect on its
loss. Moreover, by plugging the fitted familial scaling law into the definition
of Efficiency Leverage $EL(\mathcal{X}_{\mathrm{Fam}} \mid
\mathcal{X}_{\mathrm{Dense}})$, we obtain $EL$–FLOPs curves that remain
consistently above $1$ across all compute budgets considered, quantitatively
confirming that Familial Models achieve uniformly lower loss than sets of independent dense models trained under the same compute constraints. Taken together, these
findings imply a favorable architectural trade-off: Familial Models training can
amortize high pretraining costs across multiple deployment sizes with minimal
degradation in the performance of the main trunk, allowing practitioners to
obtain a spectrum of sub-models at different inference budgets from a single
training run while largely preserving the scaling behavior expected from dense
single-exit training.

\textbf{Extending to Complex Modalities and Tasks.}
This inherent flexibility holds significant promise for other domains. For instance, in Multi-Intent Spoken Language Understanding (SLU), recent surveys highlight the necessity of joint modeling to capture complex intent-slot interactions \citep{wu2025multi}. Familial Model architectures could adaptively allocate compute based on the complexity of the user's utterance, efficiently handling multi-intent scenarios with lower latency. Similarly, in the realm of multimedia security, frameworks like Aperture have demonstrated the value of patch-aware mechanisms for joint forgery detection and localization \citep{yang2025aperture}. Future work could explore integrating Familial backbones into such detection systems, allowing for rapid, coarse-grained screening at early exits and fine-grained, pixel-level localization at deeper layers.

\textbf{Enabling Collaborative Ecosystems.}
Furthermore, the ``relay-style'' inference capability of Familial Models aligns naturally with the emerging trend of multi-model collaboration \citep{10884554}. As demonstrated by recent advances in enhanced tool invocation, decoupling reasoning from format normalization via specialized collaborative models significantly improves reliability \citep{zhang2025enhanced}. Familial Models can serve as the efficient infrastructure for such agentic workflows, where shallower sub-models handle routine formatting or filtering tasks, while deeper sub-models are reserved for complex reasoning and tool selection, thereby realizing a truly ubiquitous and equitable intelligence ecosystem.

\section{Conclusion}

Based on the classic scaling law formulation, this study introduces a granularity factor $G$ to extend the scaling law specifically for Familial Models, providing a theoretical foundation for the ``train once, obtain multiple models'' paradigm. By fitting over 100 sets of experimental data, we derive the following unified formula:
\begin{equation}
L(N, D, G) = \left( 1.0059 + \frac{403.4289}{N^{0.2982}} + \frac{2980.058}{D^{0.3412}} \right) \cdot G^{0.0333}\tag{2},
\end{equation}

Experimental results indicate that the exponent $\gamma$ for granularity $G$ is extremely small ($\gamma \approx 0.0333$). This suggests that for a given model size $N$ and training token count $D$, increasing granularity $G$ incurs only a negligible penalty on the loss function. Since the value of $G^{\gamma}$ remains very close to $1$ across a wide range, the average loss of Familial Models exhibits extremely low sensitivity to variations in granularity. This characteristic offers significant advantages for engineering practice: under an equivalent compute budget, Familial Models can yield multiple sub-models of varying sizes in a single training run without significantly compromising performance, thereby adaptively meeting diverse application requirements.

To further quantify how individual branches behave relative to
size-matched dense models, we propose a branch-level scaling relation:
\[
L(P, Q, D) = L_{\mathrm{dense}} 
  + \bigl(1\times 10^{-3} \cdot P + 0.0397 \cdot Q\bigr)
    \left(\frac{2.75\times 10^{6}}{D}\right)^{0.5734}
    \tag{4},
\]
where $P$ and $Q$ denote the numbers of branch points preceding and following
the exit under consideration, respectively. This formulation explicitly
separates the penalties on loss induced by upstream and downstream branches,
with the fitted coefficients showing that additional upstream branches have
only a minimal effect on branch performance. Consequently, Familial Models
can add multiple intermediate exits without materially degrading the
performance of the main trunk, effectively yielding a family of sub-models at
different scales from a single training run.

Building on these scaling laws, we introduce an Efficiency Leverage ($EL$)
metric to capture the compute advantage of Familial Models over dense
models, defined as the ratio between their respective average losses under a
matched compute budget. Empirical evaluation using the fitted Familial Models
Scaling Law shows that $EL$ remains consistently greater than $1$ across all FLOPs
regimes, with the gain most pronounced in the low-compute setting. 

Our research demonstrates the superiority of Familial Models in engineering practice. By enabling the acquisition of multiple models of varying sizes through a single training run, this architecture effectively addresses the demand for diverse deployment scales under fixed compute budgets, while maintaining performance comparable to dense model baselines.

\bibliographystyle{plainnat}
\bibliography{paper}

\end{document}